\documentclass{article}
\usepackage{ijcai24}

\usepackage{times}
\usepackage{soul}
\usepackage{url}
\usepackage[hidelinks]{hyperref}
\usepackage[utf8]{inputenc}
\usepackage[small]{caption}
\usepackage{graphicx}
\usepackage{amsmath}
\usepackage{amsthm}
\usepackage{booktabs}
\usepackage{algorithm}
\usepackage{algorithmic}
\usepackage[switch]{lineno}

\title{Intelligent Director: An Automatic Framework for 
Dynamic Visual Composition using ChatGPT}

\author{
Sixiao Zheng\thanks{Equal contribution}\and
Jingyang Huo\footnotemark[1]\and
Yu Wang\And
Yanwei Fu\\
\affiliations
Fudan University\\
\emails
\{sxzheng18, yu\_w13, yanweifu\}@fudan.edu.cn,
jyhuo22@m.fudan.edu.cn \\
{\small \url{https://sixiaozheng.github.io/IntelligentDirector/}}
}

\begin{document}

\maketitle

\begin{abstract}
With the rise of short video platforms represented by TikTok, the trend of users expressing their creativity through photos and videos has increased dramatically. However, ordinary users lack the professional skills to produce high-quality videos using professional  creation software. To meet the demand for intelligent and user-friendly video creation tools,  we propose the Dynamic Visual Composition (DVC) task, an interesting and challenging task that aims to automatically integrate various media elements based on user requirements and create storytelling videos.  We propose an Intelligent Director framework, utilizing LENS to generate descriptions for images and video frames and combining ChatGPT to generate coherent captions while recommending appropriate music names. Then, the best-matched music is obtained through music retrieval. Then, materials such as captions, images, videos, and music are integrated to seamlessly synthesize the video. Finally, we apply AnimeGANv2 for style transfer. We construct UCF101-DVC and Personal Album datasets and verified the effectiveness of our framework in solving DVC through qualitative and quantitative comparisons, along with user studies, demonstrating its substantial potential.
\end{abstract}

\section{Introduction}
With the booming development of short video platforms such as \textit{TikTok\footnote{\url{https://www.tiktok.com/}}}, more and more users around the world use TikTok, and its Daily Active User (DAU) has exceeded one billion. A large number of users take photos and videos, and use video creation technology to upload exquisite works to the short video platform to share their wonderful creativity. Video creation involves the clever combination of caption, images, videos,  music, special effects and other materials to create expressive new videos. However, creating excellent video works requires not only artistic aesthetics, but also proficiency in various professional creation software, such as Adobe Premiere and Final Cut. For ordinary users, creating excellent video works is extremely challenging, especially for those who lack creation thinking or are not familiar with creation software. Therefore, the demand for intelligent, user-friendly video creation tools is increasing rapidly.

\begin{figure}[t]
    \centering
    \includegraphics[width=0.93\linewidth]{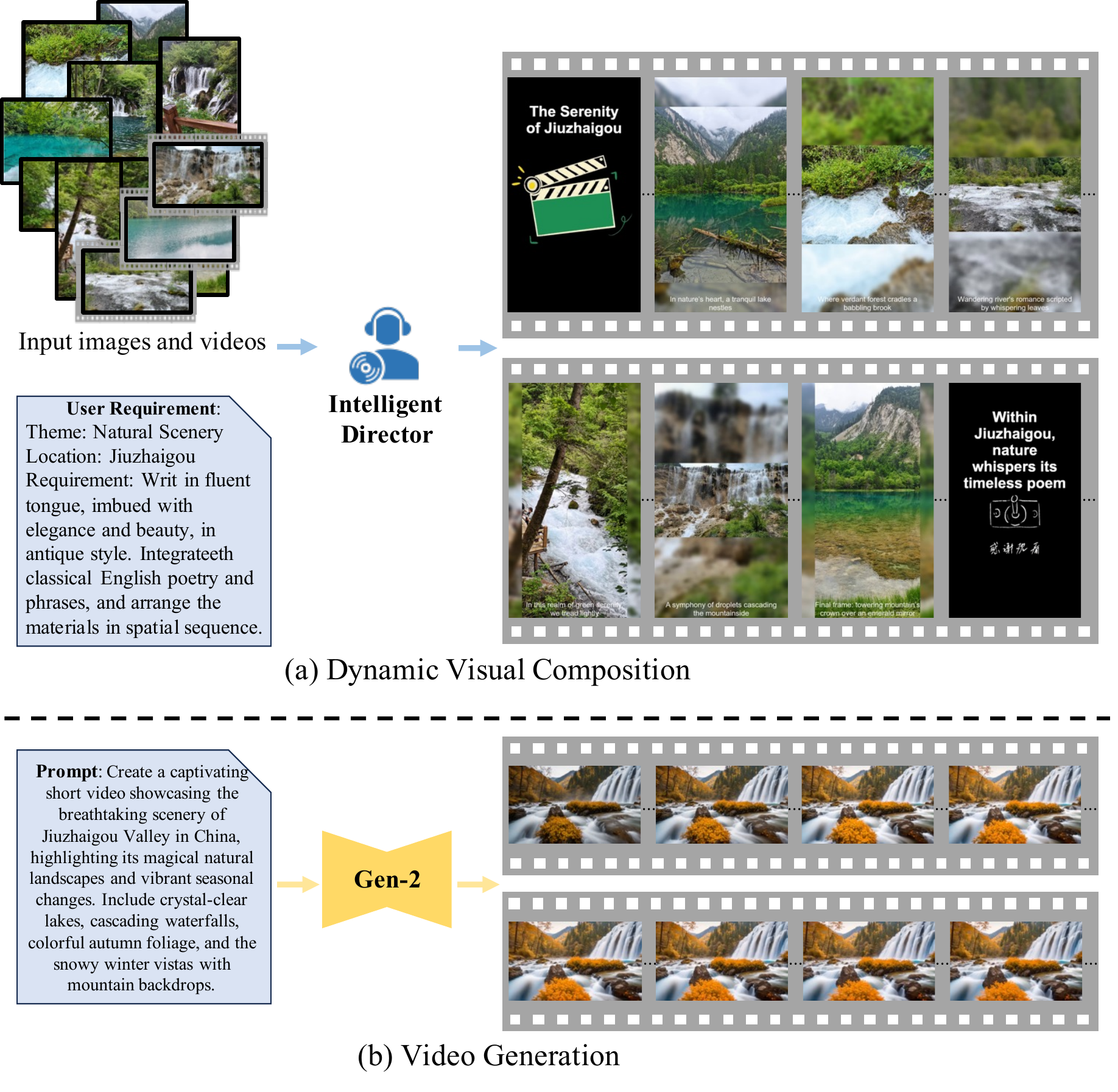}
    \caption{\textbf{Dynamic Visual Composition.} }
    \label{fig:DVC}
\end{figure}

 As shown in Figure 1(a), we formalize intelligent video creation into a new task called \textit{Dynamic Visual Composition (DVC)}, which refers to automatically integrating multiple media elements such as text, images, videos, and audio based on user instructions or requirements to create storytelling videos.
 DVC requires well alignment of visual elements with captions and music to effectively convey the story of creator  in  video format. Therefore, overall, it is a very interesting and challenging artificial intelligence task. To the best of our knowledge,  no previous AI methods can solve it well.  
 DVC can be used not only in entertainment fields such as short video platforms, but also in many fields such as advertising, education, and virtual reality. For example, in digital advertising, DVC can vividly display product features and usage scenarios to better attract the attention of users.

However, dynamic visual composition is a very challenging task that requires multiple capabilities, including 1) \textit{Multi-modal understanding.} Deeply understanding the semantics of text, images, videos, and audio to extract important information; 2) \textit{Sequencing reasoning.} Reasoning about the relationship between images and videos to form permutations and combinations that build high-quality storylines; (3)\textit{ Video Composition.} Create an artistic video by cleverly combining images, videos, text, audio, special effects, etc. 

Video generation is a task very related to DVC. 
As shown in Figure 1(b), the video generation model Gen-2\footnote{\url{https://research.runwayml.com/gen2}} generates a very realistic but single scene landscape video based on the input text, and the single scene is a problem with most current video generation models~\cite{ho2022imagen,singer2022make,esser2023structure}. Conversely, as shown in Figure (a), 
we propose the Intelligent Director framework to effectively solve DVC. The Intelligent Director can automatically integrating input images and videos according to user requirements to synthesize videos with a coherent story.


Our Intelligent Director framework consists of four main steps, namely Caption Generation, Music Retrieval, Video Composition, and Style Transfer.
In caption generation, we directly employ LENS~\cite{berrios2023language} to generate image descriptions of images. For videos, we first use perceptual hashing (pHash)~\cite{Zauner2010ImplementationAB} to identify video key frames, followed by using LENS to generate descriptions for each key frame. As the 
 descriptions generated by LENS are simplistic and lack  consideration of the contextual information from all images and videos, they are insufficient for constructing a storyline. Therefore, we leverage the generation and reasoning capabilities of ChatGPT to generate coherent and storytelling captions and recommend suitable music name.
In music retrieval, we use the music names recommended by ChatGPT to search a large music library, obtain the best-matched music to form a high-quality video.
In video composition, after gathering all the materials for video creation (including images, videos, captions, and music), we first insert captions into images and videos through caption fusion. Subsequently, we resize images and videos to the target resolution through material fine-tuning. During each transition between materials, a random switching animation is selected. Finally, music fusion is conducted, aligning material switches with the music beats through beat detection and playback time adjustments, ensuring a seamless integration of video and music. Finally, we employ AnimeGANv2 for style transfer, such as adopting the animated style of Kon Satoshi.

Our contributions in this paper are summarized as follows:
1) We propose Dynamic Visual Composition, an interesting and challenging task that involves integrating images, videos, and other materials based on user requirements to create storytelling videos.
2) We introduce Intelligent Director, an automatic framework based on ChatGPT, addressing DVC through four steps: caption generation, music retrieval, video composition, and style transfer.
3) We construct the UCF101-DVC Dataset and Personal Album Dataset for DVC, demonstrating the effectiveness of our framework in solving DVC through qualitative and quantitative comparisons, along with user studies, demonstrating its significant potential.


\section{Related Work}

\subsection{Video Generation}

 Recently, AI-Generated Content (AIGC) has received widespread attention. Text-to-video generation based on diffusion model has greatly improved the video generation effect, so it has received considerable attention \cite{ho2022imagen,singer2022make,hong2022cogvideo,khachatryan2023text2video,zhou2022large,rombach2022high,blattmann2023stable,esser2023structure}.
Notably, a recent trend is that due to the superior generative capabilities of large-scale text-to-image models, many methods attempt to transfer their knowledge and even extend to text-to-video generation~\cite{singer2022make,hong2022cogvideo}. 
Other works \cite{ho2022video,zhou2022large} utilizing Latent Diffusion Model (LDM) \cite{rombach2022high} introduced a temporal adjustment technique that enabled high-resolution video generation by introducing effective fine-tuning parameters. 
MovieFactory~\cite{zhu2023moviefactory} is a work most similar to our framework, which uses ChatGPT to extend user-provided text into detailed scripts, then utilizes a diffusion model to generate video, and finally obtain audio through retrieval models to achieve automatic movie creation.
Although the aforementioned works have made significant progress in generating high-fidelity videos, it is worth noting that DVC emphasizes the automatic fusion of multiple media elements to create videos, while video generation aims to generate realistic videos. Therefore, existing video generation works cannot effectively solve the challenge of DVC.

\subsection{Large Language Models}

\begin{figure*}[ht]
    \centering
    \includegraphics[width=\linewidth]{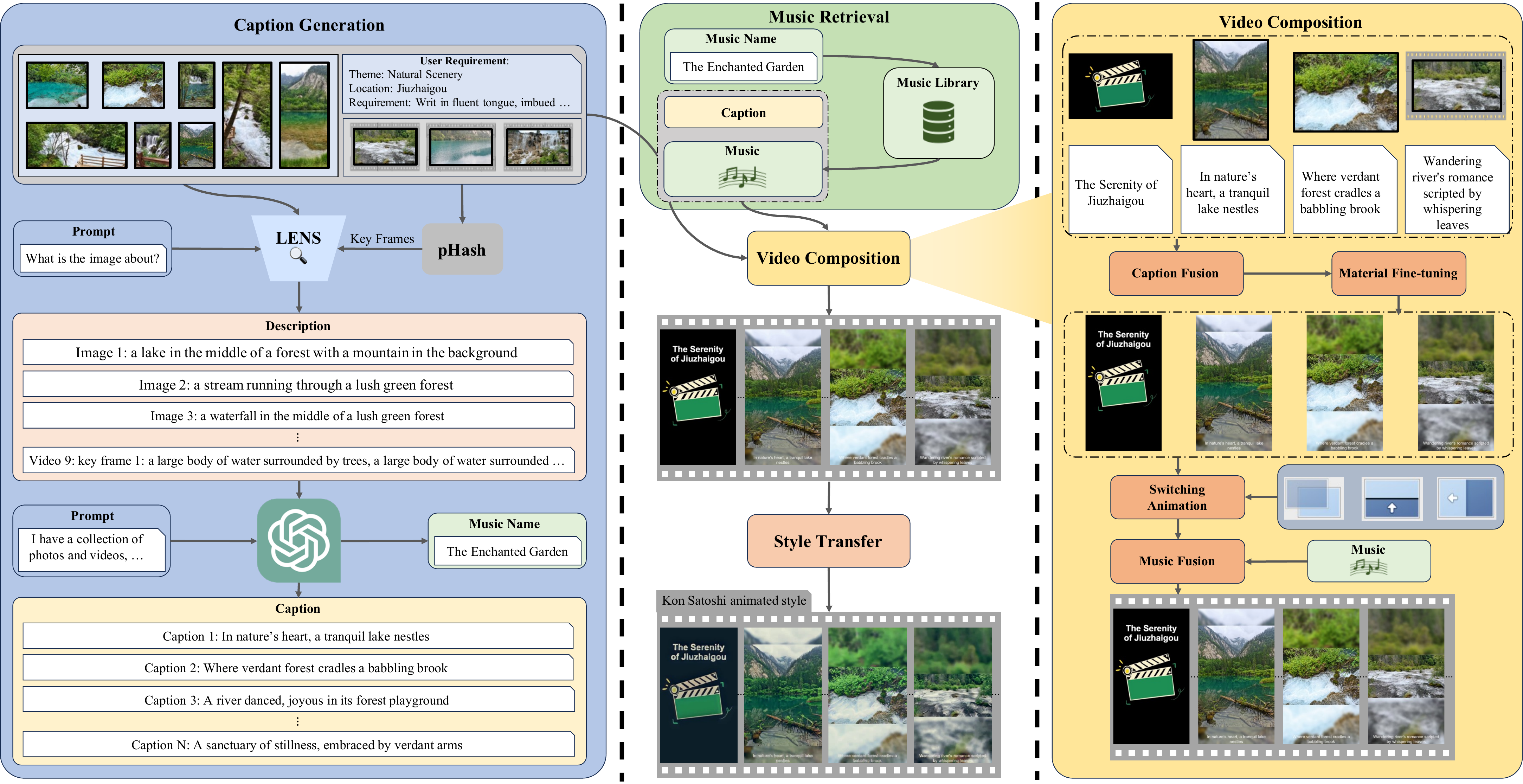}
    \caption{\textbf{An overview of our Intelligent Director framework for Dynamic Visual Composition.} Intelligent Director consists of four main steps: (1) Caption Generation, (2) Music Retrieval, (3) Video Composition, and (4) Style Transfer. In caption generation, LENS generates descriptions for images and video key frames extracted by pHash, and then ChatGPT creates coherent and storytelling captions and recommends suitable music name. In music retrieval, we utilize the music name recommended by ChatGPT to search a large music library for the best-matched music. In video composition, the seamless integration of captions, images, videos, and music is achieved through four steps: caption fusion, material fine-tuning, switching animation, and music fusion. Finally, in style transfer, AnimeGANv2 transforms the video to other styles, such as the animated style of Kon Satoshi.
    }
    \label{fig:framework}
\end{figure*}

Large language models (LLMs), characterized by their emergent abilities~\cite{wei2022emergent} and proven effectiveness across diverse complex tasks such as decision-making~\cite{li2022pre}, program synthesis~\cite{austin2021program}, and prompt engineering~\cite{zhou2022large}, are broadly classified into two categories: open source models, including LLaMA~\cite{touvron2023llama}, PaLM~\cite{chowdhery2023palm}, etc, and closed source models represented by the GPT series~\cite{radford2018improving,radford2019language,brown2020language,ouyang2022training,achiam2023gpt}. 
LLMs also perform well on tasks in other modalities~\cite{saharia2022photorealistic,koizumi2020audio,chen2023videollm,brooks2023instructpix2pix} (i.e., audio, video, and image).
Recently, there have been efforts to integrate LLMs with diffusion models~\cite{rombach2022high}, leveraging prompts generated by LLMs to produce more reliable results. 
DirecT2V \cite{hong2023large} utilizes LLMs
 to divide user inputs into separate prompts for each frame, generating frame-by-frame descriptions. These descriptions then guides diffusion model in video generation.
Free-Bloom \cite{huang2023free} employs LLMs to generate a semantic-coherence prompt sequence and LDM to generate the high-fidelity frames.
These methods combine the text processing capabilities of LLMs with the  text-to-image generation capabilities of stable diffusion. 
However, relying solely on the text processing capabilities of LLMs cannot effectively solve DVC. In this paper, we successfully combine ChatGPT with multi-modal understanding and video composition capabilities to effectively solve DVC.


\section{Overview}

\paragraph{Dynamic Visual Composition.}
In this paper, we introduce a new task called dynamic visual composition, which can automatically create storytelling videos by dynamically combining images and videos based on user requirements. 
Given a set of input images $I=\left [ I_1,I_2\dots I_n \right ] $ and videos $V=\left [ V_1,V_2\dots V_m \right ] $, along with user requirements $U$ , the goal is to create a visually and temporally coherent video $\hat{V}$ that adheres to the specified composition requirements.

DVC goes beyond a mere combination of images and videos. It requires proficiency in different fields such as image and video understanding, text understanding, image captioning, music integration, style transfer, etc. 
The main challenge is to build a fully automatic, intelligent, user-friendly framework capable of addressing  multi-modal understanding, sequencing reasoning and video composition simultaneously.

\paragraph{Intelligent Director.}
In this paper, we introduce Intelligent Director, a automatic framework for DVC. It systematically addresses the complex issues involved in generating visually engaging and temporally coherent video.
Users can start the creative process with concise descriptions, simplifying the production of video creation works. Intelligent Director adeptly tackles the challenges of democratizing video creation, providing a user-friendly tool for individuals with varying levels of expertise. 

In order to create beautiful and storytelling videos, we simulate the rule-of-thumb that directors use to create videos in practice. Initially, we rearranged and combined all the images and videos, prepare appropriate captions and music, then combine all the materials for video synthesis, and finally change the style of the video (such as animation style).

We present an overview of our Intelligent Director framework for Dynamic Visual Composition in Figure \ref{fig:framework}, 
Our framework consists of four main steps: Caption Generation, Music Retrieval, Video Composition, and Style Transfer. However, the core of our framework focuses on Caption Generation and Video Composition. 
1) In Caption Generation, we directly use LENS~\cite{berrios2023language} to generate image descriptions for images. As for videos, we initially employ Perceptual Hashing (pHash)~\cite{Zauner2010ImplementationAB} to recognize video key frames, followed by using LENS to generate descriptions for each key frame. 
Then, we utilize the generation and reasoning capabilities of ChatGPT to generate coherent and story-telling captions, along with recommending suitable music name.
2) In Music Retrieval, we leverage the music names recommended by ChatGPT to search a large music library, obtaining the best-matched music to ensure a high-quality match with the video.
3) In Video Composition, we begin by integrating captions into both images and videos through caption fusion. Following that, we adjust the scale of images and videos to the target resolution during material fine-tuning. For each transition between materials, a random switching animation is selected. Finally, music fusion is applied to synchronize the material switches with the music beats, ensuring a seamless integration of video and music.
4) In style transfer, we employ AnimeGANv2\footnote{\url{https://tachibanayoshino.github.io/AnimeGANv2/}} for style transfer, for instance, mimicking the animated style of Kon Satoshi.

\begin{figure*}[ht]
    \centering
    \includegraphics[width=0.9\linewidth]{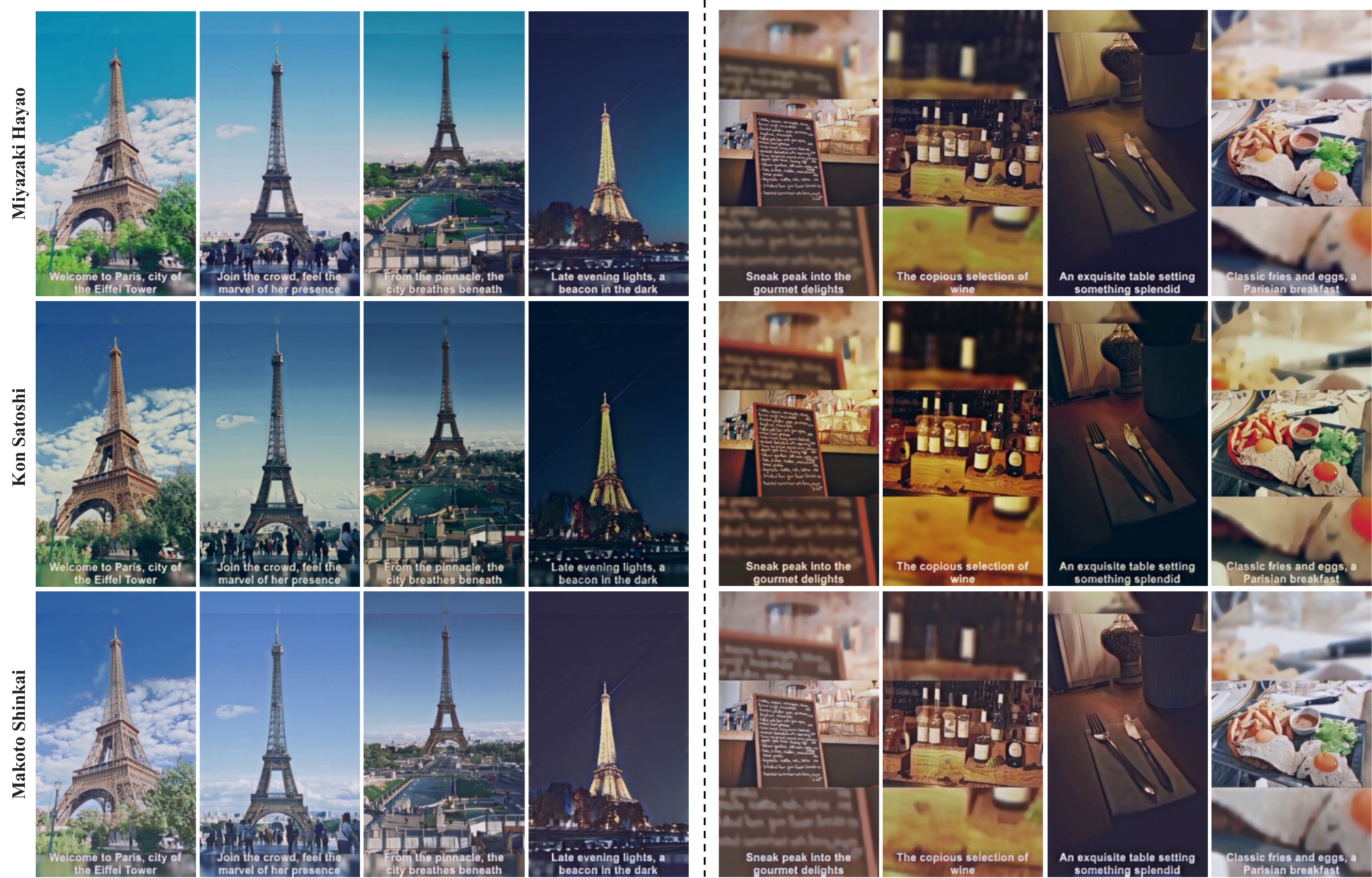}
    \caption{Results of style transfer with three animated styles on the Personal Album Dataset.}
    \label{fig:style_transfer}
\end{figure*}

\section{Method}

In this section, we first introduce the generation of captions through LENS and ChatGPT(Sec. \ref{sec:Caption_Generation}). Then, getting the best-matched music through music retrieval is introduced (Sec. \ref{sec:Music_Retrieval}). Then we introduce the four steps of Video Composition to integrate all the materials (Sec. \ref{sec:Video_Composition}). Finally, we introduce the use of AnimeGANv2 for style transfer (Sec. \ref{sec:Style_Transfer}).

\subsection{Caption Generation}
\label{sec:Caption_Generation}
LLMs have powerful text understanding and generation capabilities. We need to use the generation capabilities of LLMs to process the input of intelligent directors, including images, videos and user requirements, and then generate high-quality video captions. However, LLMs can only process text and cannot handle other modal inputs, such as images and videos. Therefore, we use the multi-modal model LENS to generate descriptions for images. LENS performs computer vision and visual reasoning tasks through a frozen LLM and a set of ``vision module''. LENS leverages these vision modules to obtain tags, attributes, and captions, which are then input into LLM to generate a response for a given question. We take the question \textit{``What is the image about''} and an image as input and obtain a description of the image.

For the video in the input of the intelligent director, we adopt a simple pHash algorithm to recognize the key frames of the video. 
If the features are similar, pHash will be ``close'' to each other. Specifically, we use the pHash to calculate the similarity between video frames, and divide the video into some small segments according to the threshold (similarity $<$ 0.6). Each small segment includes a key frame, that is, the video is down-sampled into a number of discrete key frames. After obtaining the key frames of the video, we can use the LENS to provide a description for each key frame.

The descriptions generated by the LENS model are simple descriptions of the images, lacking consideration of the contextual information from all images and videos, which are far from being story-telling captions. 
On the other hand, the initial arrangement order for the images and videos input by the user often cannot constitute a coherent story. Consequently, the initial arrangement order of the descriptions corresponding to the images and video key frames also fails to form cohesive video captions.
To address this challenge, we propose to leverage the generation and reasoning capabilities of ChatGPT to generate high-quality and coherent video captions. Specifically, we use the generation capability of ChatGPT to expand the descriptions and obtain high-quality captions describing scenes. In addition, we use the reasoning capability of ChatGPT to reason about the generated captions, so as to obtain a sequence that can form a coherent story.

We integrate user requirments through careful prompt design to ensure that the generated video captions can form a coherent story and facilitate subsequent music retrieval and video composition. Users can input requirements to control the \textit{\{theme, time, location, requirement\}} of the story. We design the prompt as: 

\noindent\textbf{\textit{prompt} = \textit{task description} + \textit{input descriptions} + \textit{detailed requirements}}.


Among them, \textit{task description} is a brief description of the task and specifies the topic, time and location (can be empty). The design of \textit{task description} is as follows:

\begin{quote} 
\textbf{task description}: ``
{\small
I have a collection of photos and videos, but their order is chaotic. I hope you can help me arrange these materials in a certain order to create a video centered around the theme \{theme\}. Additionally, I'd like you to provide a smoothly written  script that connects these images and videos into a cohesive story.
The photos and videos were taken at \{location\}.
They were captured at \{time\}
I will provide descriptions for each image or video to give you an understanding of their content.}
''
\end{quote} 

\noindent \textit{input description} are the captions generated by LENS in the previous Caption Generation. 
\textit{input description} follows the following format:

\begin{quote} 
\small
\textbf{input description}: ``
\vspace{-3.5pt}
\begin{equation*}
\begin{aligned}
&\text{Image 1: a lake in the middle of a forest with a mountain ...} \\
&\text{Image 2: a stream running through a lush green forest} \\
&\text{Image 3: a waterfall in the middle of a lush green forest}\\
\vspace{-4pt}
&~~~~~~~~~~~~~~~~~~~~~~~~~~~~~~~~~~~~~~~~~~~~~\vdots \\
\vspace{-4pt}
&\text{Video 9: key frame 1: a large body of water surrounded ...}
\end{aligned}
\end{equation*}
\vspace{-3.5pt}
''
\end{quote} %
\vspace{-3pt}

\noindent\textit{detailed requirements} are the detailed requirements of the task, including task splitting, output format, output language, word count, title recommendation, conclusion recommendation and music recommendation. The design of \textit{detailed requirements} is as follows:

\begin{quote} 
\textbf{detailed requirements}: ``
{\small
I need you to do two things:
        (1) Rearrange the materials, grouping similar images together. If there's a clear timeline, arrange them in chronological order, otherwise, organize them based on your logical sequence.\\
        (2) Write a script according to the adjusted material sequence. I hope your script meets the following requirements: \{requirement\}. It should be concise, fluent, vivid, and the transitions between different materials should be natural. Each caption for the materials should not exceed 20 words. ...
        }''
\end{quote} 

By structuring prompts in this way, we can leverage ChatGPT to generate high-quality and coherent video captions with recommended titles, conclusion, and music. Due to space limitations, readers are encouraged to refer to the supplementary materials for the complete prompt.




\subsection{Music Retrieval}
\label{sec:Music_Retrieval}
Music plays an irreplaceable role in video creation. Its selection is not only to fill gaps, but also to enhance the audience's perceptual experience. 
During the caption generation, we cleverly used prompts to ask ChatGPT to recommend music name that match the video. 
We adopt a retrieval-based approach by searching a large music library to obtain the best-matched music to ensure that the selected music complements the overall presentation of the video.

\subsection{Video Composition}
\label{sec:Video_Composition}
After acquiring materials for video creation, including images, videos, captions, and music, the next step is to combine all materials to create the synthesized video, referred to as video composition. We break down video composition into the following four steps:


\paragraph{Caption Fusion} 
We first created a template for the opening and ending, inserted the title generated by ChatGPT into the opening, and inserted the conclusion into the ending. At the same time, insert the scripts generated by ChatGPT into the corresponding images or videos in order.

\paragraph{Material Fine-tuning} 
Due to variations in resolution and aspect ratio among materials, adjustments are necessary to ensure uniform sizing. Specifically, assuming the original height and width of a material are $H_O$ and $W_O$
  respectively, and the target height and width (i.e., the dimensions of the synthesized video) are $H_T$ and $W_T$, we resize the material to two scales: $(H_T, \frac{H_T}{H_O}W_O )$ and $(\frac{W_T}{W_O}H_O, W_T)$, denoted as $M_1$ and $M_2$. If $\frac{H_T}{H_O}W_O\le W_T$, $M_1$ is centered, and $M_2$ is Gaussian blurred and placed beneath to prevent black regions around $M_1$. Conversely, if $\frac{W_T}{W_O}H_O\le H_T$, $M_2$ is centered, and $M_1$ is  Gaussian blurred and placed beneath.

\paragraph{Switching Animation}
The switching animation between materials in the video provides visual transitions and maintains coherence.
We create four switching animations for material switching, including crossfade in, crossfade out, upward translation, and lateral translation. Random selection is made for each material switching.

\begin{figure*}[ht]
    \centering
    \includegraphics[width=0.9\linewidth]{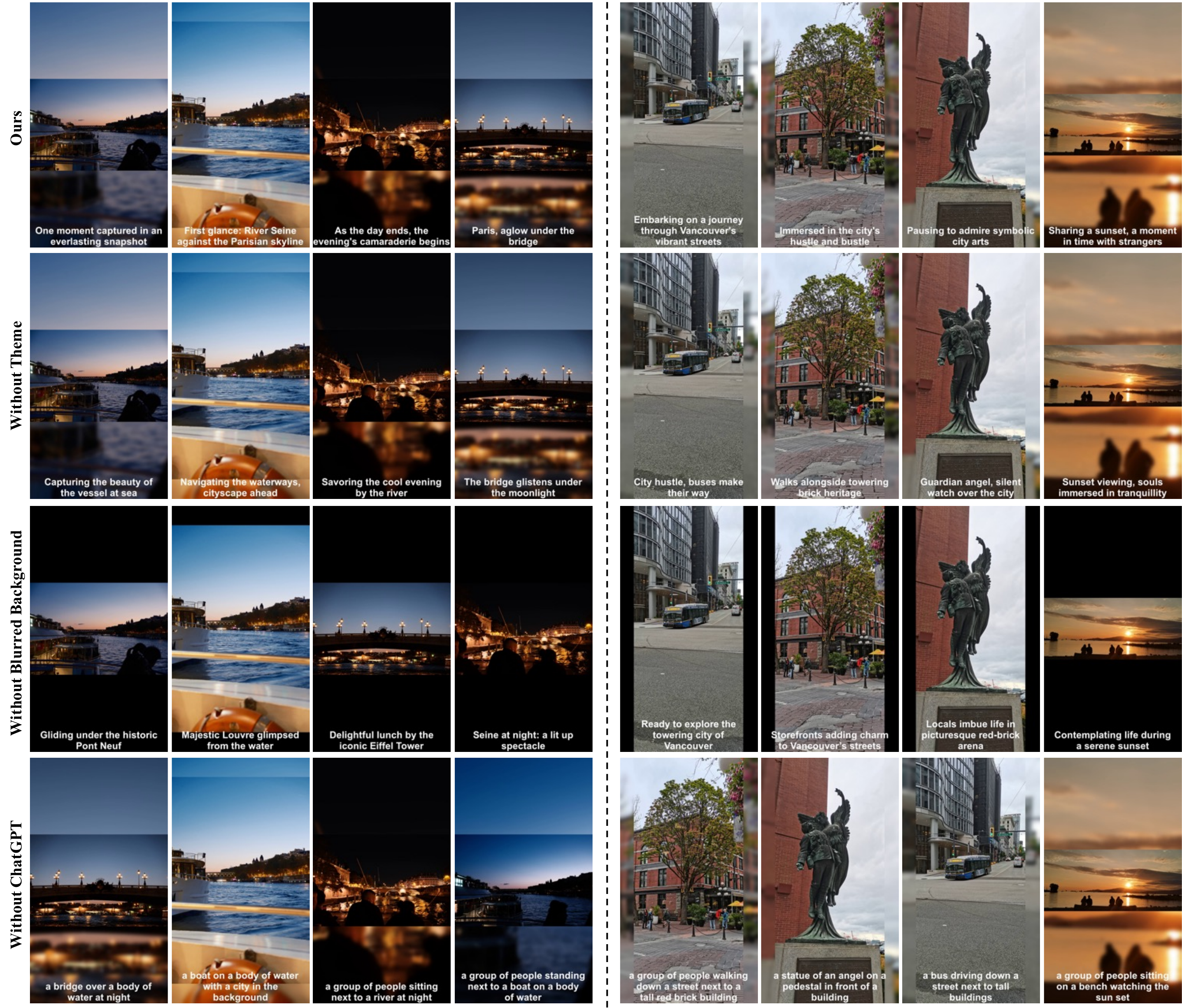}
    \caption{Comparison results between our framework and three baseline models on the Personal Album Dataset.}
    \label{fig:Res_PAD}
\end{figure*}

\paragraph{Music Fusion}
During music fusion, standard beat detection~\cite{ellis2007beat} is initially performed, followed by fine-tuning the duration of each material based on the music's rhythm. Specifically, setting the duration of the image defaults to four seconds. To align each material with the beat of the music when switching 
, the beat moment closest to the end time of the material will be identified as the new end time.
Then, fine-tune the duration of the image or video to match the beat. 
Additionally, a minimum duration is set for each image or video clip to prevent flickering during video playback. This carefully designed process is designed to ensure seamless integration of music and video.


\subsection{Style Transfer}
\label{sec:Style_Transfer}
Video style transfer empowers users with creative freedom by transforming the style of a video, making the content more aligned with specific themes.
We utilize AnimeGANv2~\cite{animeganv2} for style transfer on the generated videos, offering support for three animated styles: Miyazaki Hayao, Makoto Shinkai, and Kon Satoshi. Firstly, we extract the audio from the video, decompose the input video into individual frames, and subsequently apply AnimeGANv2 for style transfer on each video frame. Finally, we merge the audio with the style-transferred video frames to create a new video. 

\section{Experiments}

\begin{figure}[t]
    \centering
    \includegraphics[width=\linewidth]{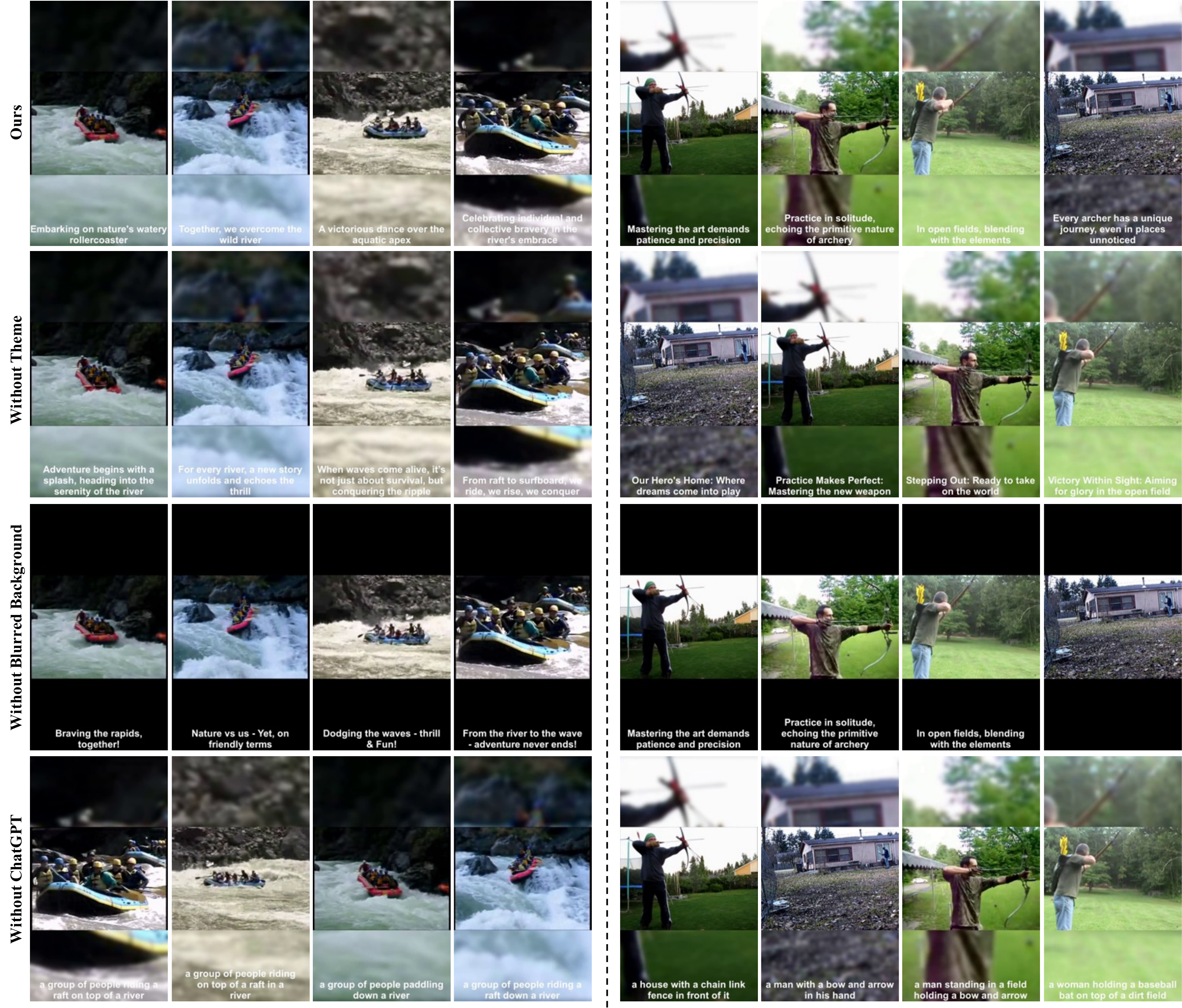}
    \caption{Comparison results between our framework and three baseline models on the UCF101-DVC Dataset.}
    \label{fig:Res_UCF101}
    \vspace{-3pt}
\end{figure}

\subsection{Datasets}
The DVC task has not been specifically studied before, so we construct UCF101-DVC Dataset and Personal Album Dataset to verify the effectiveness of our proposed framework.
\paragraph{UCF101-DVC Dataset}
UCF101 dataset~\cite{soomro2012ucf101} consists of 13,320 video clips, which are classified into 101 categories.  We randomly sampled 8 video clips from each class to form a video set, resulting in a total of 101 video sets, encompassing 808 video clips.

\paragraph{Personal Album Dataset}
This dataset consists of photos and videos captured by five volunteers using their smartphones or cameras, totaling 25 sets of data, including 195 photos and 8 videos. The number of photos and videos in each set varies, with a maximum of 13 photos and 3 videos. Each set of data was captured at a specific location or centered around a particular theme, annotated by the volunteers.

\subsection{Evaluation Metrics}
We use Type-Token Ratio (TTR) to evaluate the diversity of the generated video scripts. TTR measures the text diversity by calculating the ratio of the number of different words to the total number of words in a text. We compute the TTR for each video captions individually, and then average these values to obtain the overall mean TTR of all the videos. A higher TTR indicates increased diversity in the text.


\subsection{Implementation details}
In the caption generation, we utilize the image-match package to invoke the pHash algorithm, setting the similarity threshold of $<$ 0.6 to extract keyframes from videos. Subsequently, we employ the open-source code from LENS to generate descriptions. 
In the music retrieval, we use 9 music library, including QQMusic, NetEase Cloud Music, Kugou Music, KuWo Music, QianQian music, 5sing music, Migu Music, JOOX Music, 1ting music.
In the video composition, we leverage the moviepy package to perform various video operations. The target resolution $(H_T \times  W_T)$ for videos is set at $1280 \times 720$ (720P) or $1920 \times 1080$ (1080P). 
For music fusion, we employ the librosa package to detect beats in the music. In the style transfer, we utilize the open-source code of AnimeGANv2 to perform style transfer on videos.

\subsection{Baselines}
To evaluate the effectiveness of our Intelligent Director framework, we construct three baseline models for comparison: 
1) \textbf{Without Theme}: The theme in user requirements significantly affects the quality of captions generated by ChatGPT. We thus remove the theme to analyze its impact on results.
2) \textbf{Without Blurred Background}: Gaussian blur in Material Fine-tuning was excluded to study its influence on overall visual aesthetics.
3) \textbf{Without ChatGPT}: Replacing ChatGPT with a method randomly ordering descriptions to examine its contribution to caption quality.

\subsection{Qualitative Results}

As shown in Figure \ref{fig:style_transfer}, we display key frames extracted from the generated videos. The captions generated by our framework are closely related to the videos, forming a coherent story, and the visuals become more exquisite after style transfer to three animation styles. As shown in Figure \ref{fig:Res_PAD}, \ref{fig:Res_UCF101}, the results of ``Without Theme'' indicate that user-input themes can make captions more fitting to the visual effects of the videos. The results of ``Without Blurred Background'' suggest that incorporating a Gaussian-blurred background can significantly enhance the aesthetics of the videos. The results of ``Without ChatGPT'' demonstrate that ChatGPT can generate more coherent and storytelling captions.

\subsection{Quantitative Results}

We report the TTR metric in Table \ref{tab:metric} to compare the text diversity of our framework and three baselines. Our framework, incorporating ChatGPT, notably outperforms the baseline without ChatGPT. This indicates that the text generated by the LENS model includes a significant amount of repetition. Importantly, in tasks like DVC, this repetitive and low-quality text can negatively impact the user experience. Additionally, our framework shows a slight decrease in performance compared to the baseline without a theme. One potential reason is that a theme-centered script tends to focus more tightly on a central idea, resulting in slightly less diversity.

\subsection{User Study}


We conduct a comprehensive user study involving four methods, including our Intelligent Director framework and three baseline variations derived from the framework. Each method is evaluated on a set of 10 videos. A total of 15 participants take part in the study, providing ratings based on four dimensions: 1) Correspondence: the alignment between the captions and the video; 2) Coherence: the coherence and vividness of the captions; 3) Matching: the Matching between music and video; 4) 
Quality: the overall quality of the video.
All evaluations are conducted using a Likert scale ranging from 1 to 5. As shown in Table \ref{tab:user_study}, our framework outperforms three baseline models in all four metrics, which indicates that our
framework is more favored by users. 
\begin{table}
\centering
\footnotesize
\setlength{\tabcolsep}{1.5mm}
\small{
\begin{tabular}{cccccc} 
\toprule 
\multicolumn{1}{c}{\textsc{Dataset}} & w/o Theme & w/o Blurred BG. & w/o ChatGPT & Ours \tabularnewline 
\midrule
UCF101-DVC & \textbf{0.813} & 0.798 & 0.259 & 0.805 \tabularnewline 
PAD & \textbf{0.824} & 0.790 & 0.440 & 0.820 \tabularnewline 
\bottomrule
\end{tabular}}
\caption{Quantitative comparison between our framework and three baseline models on TTR.} \label{tab:metric}
\vspace{-0.05in}
\end{table}

\begin{table}
\centering
\footnotesize
\setlength{\tabcolsep}{1.5pt}
\small{
\begin{tabular}{ccccc} 
\toprule 
\multicolumn{1}{c}{\textsc{Method}} & w/o Theme & w/o Blurred BG. & w/o ChatGPT & Ours \tabularnewline 
\midrule 
Correspondence & 3.73 & 3.76 & 3.42 & \textbf{3.87} \tabularnewline 
Coherence & 3.66  & 3.47 & 2.67 & \textbf{3.68}  \tabularnewline 
Matching  & 3.39 & 3.13 & 2.76 & \textbf{3.66} \tabularnewline 
Quality & 3.47 & 3.21 & 2.68 & \textbf{3.65}  \tabularnewline 
\bottomrule
\end{tabular}}
\caption{User Study Results. In this table, larger numbers indicate better performance or preference.} \label{tab:user_study}
\vspace{-0.15in}
\end{table}

\section{Conclusions}

In this paper, we introduce the Dynamic Visual Composition (DVC) task, aiming to automatically integrate various media elements based on user requirements to create storytelling videos. We propose the Intelligent Director framework, effectively addressing the DVC task through four steps: Caption Generation, Music Retrieval, Video Composition, and Style Transfer. By constructing the UCF101-DVC and Personal Album datasets and conducting qualitative and quantitative comparisons and user studies, we validate the effectiveness of the Intelligent Director framework in solving the DVC task and demonstrate its significant potential. 

\bibliographystyle{named}
\bibliography{ijcai24}

\begin{thebibliography}{}

\bibitem[\protect\citeauthoryear{Achiam \bgroup \em et al.\egroup }{2023}]{achiam2023gpt}
Josh Achiam, Steven Adler, Sandhini Agarwal, Lama Ahmad, Ilge Akkaya, Florencia~Leoni Aleman, Diogo Almeida, Janko Altenschmidt, Sam Altman, Shyamal Anadkat, et~al.
\newblock Gpt-4 technical report.
\newblock {\em arXiv preprint arXiv:2303.08774}, 2023.

\bibitem[\protect\citeauthoryear{Austin \bgroup \em et al.\egroup }{2021}]{austin2021program}
Jacob Austin, Augustus Odena, Maxwell Nye, Maarten Bosma, Henryk Michalewski, David Dohan, Ellen Jiang, Carrie Cai, Michael Terry, Quoc Le, et~al.
\newblock Program synthesis with large language models.
\newblock {\em arXiv preprint arXiv:2108.07732}, 2021.

\bibitem[\protect\citeauthoryear{Berrios \bgroup \em et al.\egroup }{2023}]{berrios2023language}
William Berrios, Gautam Mittal, Tristan Thrush, Douwe Kiela, and Amanpreet Singh.
\newblock Towards language models that can see: Computer vision through the lens of natural language, 2023.

\bibitem[\protect\citeauthoryear{Blattmann \bgroup \em et al.\egroup }{2023}]{blattmann2023stable}
Andreas Blattmann, Tim Dockhorn, Sumith Kulal, Daniel Mendelevitch, Maciej Kilian, Dominik Lorenz, Yam Levi, Zion English, Vikram Voleti, Adam Letts, et~al.
\newblock Stable video diffusion: Scaling latent video diffusion models to large datasets.
\newblock {\em arXiv preprint arXiv:2311.15127}, 2023.

\bibitem[\protect\citeauthoryear{Brooks \bgroup \em et al.\egroup }{2023}]{brooks2023instructpix2pix}
Tim Brooks, Aleksander Holynski, and Alexei~A Efros.
\newblock Instructpix2pix: Learning to follow image editing instructions.
\newblock In {\em Proceedings of the IEEE/CVF Conference on Computer Vision and Pattern Recognition}, pages 18392--18402, 2023.

\bibitem[\protect\citeauthoryear{Brown \bgroup \em et al.\egroup }{2020}]{brown2020language}
Tom Brown, Benjamin Mann, Nick Ryder, Melanie Subbiah, Jared~D Kaplan, Prafulla Dhariwal, Arvind Neelakantan, Pranav Shyam, Girish Sastry, Amanda Askell, et~al.
\newblock Language models are few-shot learners.
\newblock {\em Advances in neural information processing systems}, 33:1877--1901, 2020.

\bibitem[\protect\citeauthoryear{Chen and Liu}{2021}]{animeganv2}
Xin Chen and Gang Liu.
\newblock Animeganv2, 2021.

\bibitem[\protect\citeauthoryear{Chen \bgroup \em et al.\egroup }{2023}]{chen2023videollm}
Guo Chen, Yin-Dong Zheng, Jiahao Wang, Jilan Xu, Yifei Huang, Junting Pan, Yi~Wang, Yali Wang, Yu~Qiao, Tong Lu, et~al.
\newblock Videollm: Modeling video sequence with large language models.
\newblock {\em arXiv preprint arXiv:2305.13292}, 2023.

\bibitem[\protect\citeauthoryear{Chowdhery \bgroup \em et al.\egroup }{2023}]{chowdhery2023palm}
Aakanksha Chowdhery, Sharan Narang, Jacob Devlin, Maarten Bosma, Gaurav Mishra, Adam Roberts, Paul Barham, Hyung~Won Chung, Charles Sutton, Sebastian Gehrmann, et~al.
\newblock Palm: Scaling language modeling with pathways.
\newblock {\em Journal of Machine Learning Research}, 24(240):1--113, 2023.

\bibitem[\protect\citeauthoryear{Ellis}{2007}]{ellis2007beat}
Daniel~PW Ellis.
\newblock Beat tracking by dynamic programming.
\newblock {\em Journal of New Music Research}, 36(1):51--60, 2007.

\bibitem[\protect\citeauthoryear{Esser \bgroup \em et al.\egroup }{2023}]{esser2023structure}
Patrick Esser, Johnathan Chiu, Parmida Atighehchian, Jonathan Granskog, and Anastasis Germanidis.
\newblock Structure and content-guided video synthesis with diffusion models.
\newblock In {\em Proceedings of the IEEE/CVF International Conference on Computer Vision}, pages 7346--7356, 2023.

\bibitem[\protect\citeauthoryear{Ho \bgroup \em et al.\egroup }{2022a}]{ho2022imagen}
Jonathan Ho, William Chan, Chitwan Saharia, Jay Whang, Ruiqi Gao, Alexey Gritsenko, Diederik~P Kingma, Ben Poole, Mohammad Norouzi, David~J Fleet, et~al.
\newblock Imagen video: High definition video generation with diffusion models.
\newblock {\em arXiv preprint arXiv:2210.02303}, 2022.

\bibitem[\protect\citeauthoryear{Ho \bgroup \em et al.\egroup }{2022b}]{ho2022video}
Jonathan Ho, Tim Salimans, Alexey Gritsenko, William Chan, Mohammad Norouzi, and David~J Fleet.
\newblock Video diffusion models.
\newblock {\em arXiv:2204.03458}, 2022.

\bibitem[\protect\citeauthoryear{Hong \bgroup \em et al.\egroup }{2022}]{hong2022cogvideo}
Wenyi Hong, Ming Ding, Wendi Zheng, Xinghan Liu, and Jie Tang.
\newblock Cogvideo: Large-scale pretraining for text-to-video generation via transformers.
\newblock {\em arXiv preprint arXiv:2205.15868}, 2022.

\bibitem[\protect\citeauthoryear{Hong \bgroup \em et al.\egroup }{2023}]{hong2023large}
Susung Hong, Junyoung Seo, Sunghwan Hong, Heeseong Shin, and Seungryong Kim.
\newblock Large language models are frame-level directors for zero-shot text-to-video generation.
\newblock {\em arXiv preprint arXiv:2305.14330}, 2023.

\bibitem[\protect\citeauthoryear{Huang \bgroup \em et al.\egroup }{2023}]{huang2023free}
Hanzhuo Huang, Yufan Feng, Cheng Shi, Lan Xu, Jingyi Yu, and Sibei Yang.
\newblock Free-bloom: Zero-shot text-to-video generator with llm director and ldm animator.
\newblock {\em arXiv preprint arXiv:2309.14494}, 2023.

\bibitem[\protect\citeauthoryear{Khachatryan \bgroup \em et al.\egroup }{2023}]{khachatryan2023text2video}
Levon Khachatryan, Andranik Movsisyan, Vahram Tadevosyan, Roberto Henschel, Zhangyang Wang, Shant Navasardyan, and Humphrey Shi.
\newblock Text2video-zero: Text-to-image diffusion models are zero-shot video generators.
\newblock {\em arXiv preprint arXiv:2303.13439}, 2023.

\bibitem[\protect\citeauthoryear{Koizumi \bgroup \em et al.\egroup }{2020}]{koizumi2020audio}
Yuma Koizumi, Yasunori Ohishi, Daisuke Niizumi, Daiki Takeuchi, and Masahiro Yasuda.
\newblock Audio captioning using pre-trained large-scale language model guided by audio-based similar caption retrieval.
\newblock {\em arXiv preprint arXiv:2012.07331}, 2020.

\bibitem[\protect\citeauthoryear{Li \bgroup \em et al.\egroup }{2022}]{li2022pre}
Shuang Li, Xavier Puig, Chris Paxton, Yilun Du, Clinton Wang, Linxi Fan, Tao Chen, De-An Huang, Ekin Aky{\"u}rek, Anima Anandkumar, et~al.
\newblock Pre-trained language models for interactive decision-making.
\newblock {\em Advances in Neural Information Processing Systems}, 35:31199--31212, 2022.

\bibitem[\protect\citeauthoryear{Ouyang \bgroup \em et al.\egroup }{2022}]{ouyang2022training}
Long Ouyang, Jeffrey Wu, Xu~Jiang, Diogo Almeida, Carroll Wainwright, Pamela Mishkin, Chong Zhang, Sandhini Agarwal, Katarina Slama, Alex Ray, et~al.
\newblock Training language models to follow instructions with human feedback.
\newblock {\em Advances in Neural Information Processing Systems}, 35:27730--27744, 2022.

\bibitem[\protect\citeauthoryear{Radford \bgroup \em et al.\egroup }{2018}]{radford2018improving}
Alec Radford, Karthik Narasimhan, Tim Salimans, Ilya Sutskever, et~al.
\newblock Improving language understanding by generative pre-training.
\newblock 2018.

\bibitem[\protect\citeauthoryear{Radford \bgroup \em et al.\egroup }{2019}]{radford2019language}
Alec Radford, Jeffrey Wu, Rewon Child, David Luan, Dario Amodei, Ilya Sutskever, et~al.
\newblock Language models are unsupervised multitask learners.
\newblock {\em OpenAI blog}, 1(8):9, 2019.

\bibitem[\protect\citeauthoryear{Rombach \bgroup \em et al.\egroup }{2022}]{rombach2022high}
Robin Rombach, Andreas Blattmann, Dominik Lorenz, Patrick Esser, and Bj{\"o}rn Ommer.
\newblock High-resolution image synthesis with latent diffusion models.
\newblock In {\em Proceedings of the IEEE/CVF conference on computer vision and pattern recognition}, pages 10684--10695, 2022.

\bibitem[\protect\citeauthoryear{Saharia \bgroup \em et al.\egroup }{2022}]{saharia2022photorealistic}
Chitwan Saharia, William Chan, Saurabh Saxena, Lala Li, Jay Whang, Emily~L Denton, Kamyar Ghasemipour, Raphael Gontijo~Lopes, Burcu Karagol~Ayan, Tim Salimans, et~al.
\newblock Photorealistic text-to-image diffusion models with deep language understanding.
\newblock {\em Advances in Neural Information Processing Systems}, 35:36479--36494, 2022.

\bibitem[\protect\citeauthoryear{Singer \bgroup \em et al.\egroup }{2022}]{singer2022make}
Uriel Singer, Adam Polyak, Thomas Hayes, Xi~Yin, Jie An, Songyang Zhang, Qiyuan Hu, Harry Yang, Oron Ashual, Oran Gafni, et~al.
\newblock Make-a-video: Text-to-video generation without text-video data.
\newblock {\em arXiv preprint arXiv:2209.14792}, 2022.

\bibitem[\protect\citeauthoryear{Soomro \bgroup \em et al.\egroup }{2012}]{soomro2012ucf101}
Khurram Soomro, Amir~Roshan Zamir, and Mubarak Shah.
\newblock Ucf101: A dataset of 101 human actions classes from videos in the wild.
\newblock {\em arXiv preprint arXiv:1212.0402}, 2012.

\bibitem[\protect\citeauthoryear{Touvron \bgroup \em et al.\egroup }{2023}]{touvron2023llama}
Hugo Touvron, Thibaut Lavril, Gautier Izacard, Xavier Martinet, Marie-Anne Lachaux, Timoth{\'e}e Lacroix, Baptiste Rozi{\`e}re, Naman Goyal, Eric Hambro, Faisal Azhar, et~al.
\newblock Llama: Open and efficient foundation language models.
\newblock {\em arXiv preprint arXiv:2302.13971}, 2023.

\bibitem[\protect\citeauthoryear{Wei \bgroup \em et al.\egroup }{2022}]{wei2022emergent}
Jason Wei, Yi~Tay, Rishi Bommasani, Colin Raffel, Barret Zoph, Sebastian Borgeaud, Dani Yogatama, Maarten Bosma, Denny Zhou, Donald Metzler, et~al.
\newblock Emergent abilities of large language models.
\newblock {\em arXiv preprint arXiv:2206.07682}, 2022.

\bibitem[\protect\citeauthoryear{Zauner}{2010}]{Zauner2010ImplementationAB}
Christoph Zauner.
\newblock Implementation and benchmarking of perceptual image hash functions.
\newblock 2010.

\bibitem[\protect\citeauthoryear{Zhou \bgroup \em et al.\egroup }{2022}]{zhou2022large}
Yongchao Zhou, Andrei~Ioan Muresanu, Ziwen Han, Keiran Paster, Silviu Pitis, Harris Chan, and Jimmy Ba.
\newblock Large language models are human-level prompt engineers.
\newblock {\em arXiv preprint arXiv:2211.01910}, 2022.

\bibitem[\protect\citeauthoryear{Zhu \bgroup \em et al.\egroup }{2023}]{zhu2023moviefactory}
Junchen Zhu, Huan Yang, Huiguo He, Wenjing Wang, Zixi Tuo, Wen-Huang Cheng, Lianli Gao, Jingkuan Song, and Jianlong Fu.
\newblock Moviefactory: Automatic movie creation from text using large generative models for language and images.
\newblock {\em arXiv preprint arXiv:2306.07257}, 2023.

\end{thebibliography}

\clearpage
\appendix

\section*{Appendix}
\section{Complete Prompt for ChatGPT}

In the Caption Generation of Intelligent Director, we leverage the generation and inference capabilities of ChatGPT to generate high-quality and coherent video captions. We incorporate user requirements into the prompt design to ensure that the generated video captions construct a cohesive story, enabling seamless music retrieval and video composition. Users have the flexibility to input specific requirements to control the \textit{\{theme, time, location, requirement\}} of the story. Our prompt is structured as:

\noindent\textbf{\textit{prompt} = \textit{task description} + \textit{input descriptions} + \textit{detailed requirements}}.


Among them, \textit{task description} is a brief description of the task and specifies the topic, time and location (can be empty). The design of \textit{task description} is as follows:

\begin{quote} 
\textbf{task description}: ``
{\small
I have a collection of photos and videos, but their order is chaotic. I hope you can help me arrange these materials in a certain order to create a video centered around the theme \{theme\}. Additionally, I'd like you to provide a smoothly written  script that connects these images and videos into a cohesive story.
The photos and videos were taken at \{location\}.
They were captured at \{time\}
I will provide descriptions for each image or video to give you an understanding of their content.}
''
\end{quote} 

\noindent \textit{input description} are the captions generated by LENS in the previous Caption Generation. 
\textit{input description} follows the following format:

\begin{quote} 
\small
\textbf{input description}: ``
\vspace{-3.5pt}
\begin{equation*}
\begin{aligned}
&\text{Image 1: a lake in the middle of a forest with a mountain ...} \\
&\text{Image 2: a stream running through a lush green forest} \\
&\text{Image 3: a waterfall in the middle of a lush green forest}\\
\vspace{-4pt}
&~~~~~~~~~~~~~~~~~~~~~~~~~~~~~~~~~~~~~~~~~~~~~\vdots \\
\vspace{-4pt}
&\text{Video 9: key frame 1: a large body of water surrounded ...}
\end{aligned}
\end{equation*}
\vspace{-3.5pt}
''
\end{quote} %
\vspace{-3pt}

\noindent\textit{detailed requirements} are the detailed requirements of the task, including task splitting, output format, output language, word count, title recommendation, conclusion recommendation and music recommendation. The design of \textit{detailed requirements} is as follows:

\begin{quote} 
\textbf{detailed requirements}: ``
{\small
I need you to do two things:
        (1) Rearrange the materials, grouping similar images together. If there's a clear timeline, arrange them in chronological order, otherwise, organize them based on your logical sequence.\\
        (2) Write a script according to the adjusted material sequence. I hope your script meets the following requirements: \{requirement\}. It should be concise, fluent, vivid, and the transitions between different materials should be natural. Each caption for the materials should not exceed 20 words.\\
        Also, please recommend a piece of instrumental music that suits this video.
        Finally, you should first rearrange the materials and then write corresponding captions based on the rearranged sequence. Below is an example output format:\\
        Order: (A sequence of Arabic numbers separated by commas, indicating the adjusted order of materials in your script)\\
        Title: A title for the beginning of the video, not exceeding 5 words\\
        Materials: Content of the materials rearranged in order\\
        Captions: A specific Arabic number (indicating the corresponding section of the material): The specific content of the caption\\
        Closing: A closing statement at the end of the video, not exceeding 8 words\\
        Music Recommendation: (Only provide the name of the music, no other words)
        }''
\end{quote} 

By organizing prompts in this way, we can utilize ChatGPT to generate video captions of superior quality and coherence, complete with recommended titles, conclusions, and music recommendations.

\section{Additional Quantitative Results}
In the Caption Generation, we leverage the generative and reasoning capabilities of ChatGPT to produce high-quality and coherent video captions. To comprehensively evaluate the performance of our framework, we introduce GPT-4  as an evaluator. We design a specific evaluation prompt instructing GPT-4 to simulate an impartial judge in assessing the quality of  text and visual elements of the edited video. Specifically, GPT-4 is tasked with scoring each aspect, including consistency, logicality, vividness, and overall effectiveness of the text and visuals, on a scale from 1 to 5. Additionally, it is required to provide corresponding justifications for each aspect. The entire process aims to simulate a real evaluation scenario, validating the performance of our proposed framework in the Dynamic Visual Composition task.

\begin{quote} 
\textbf{GPT-4 prompt}: ``
{\small
    You are an impartial judge tasked with evaluating the quality of edited video based on textual and visual elements. Your assessment should consider the overall coherence, creativity, and effectiveness of the content. 
    You will rate the quality of the output on multiple aspects such as Consistency of text and video, Logicality, Vividness, and Overall.

    Evaluate

    Aspects 
    
    Consistency of text and video: Rate the Consistency of text and video on how well the text aligns with the visuals in the video clip, according to the consistency between what is described in the text and what is presented visually. A score of 5 indicates complete alignment, while a score of 1 suggests significant inconsistency.
    
    Logicality: Evaluate the logical flow of the text, examining how it contributes to a cohesive and sensible storyline. A score of 5 indicates a text that is logically sound, while a score of 1 suggests a lack of coherence and logic. 

    Vividness: Rate the Vividness on how well the text brings the video to life and enhances the viewer's experience. A score of 5 indicates highly vivid text, while a score of 1 suggests a lack of vividness and engagement.
    
    
    Overall: Rate the overall assessment on how effectively the text and visuals work together to create a compelling and coherent story. A score of 5 indicates good integration, while a score of 1 suggests poor integration.
    
    Format 
    Please rate the quality of the edited video by scoring it from 1 to 5 individually on each aspect. \\
    - 1: strongly disagree \\
    - 2: disagree \\
    - 3: neutral \\
    - 4: agree \\
    - 5: strongly agree \\
    
    Now, please output your scores and a short rationale below in a json format by filling in the placeholders in []:
    
    \{ \\
        "consistency of text and video": \{ \\
            "reason": "[your rationale]", \\
            "score": "[score from 1 to 5]" \\
        \},\\
        "logicality": \{ \\
            "reason": "[your rationale]",\\
            "score": "[score from 1 to 5]"\\
        \}, \\
        "vividness": \{\\
            "reason": "[your rationale]",\\
            "score": "[score from 1 to 5]"\\
        \},\\
        "aesthetic": \{\\
            "reason": "[your rationale]",\\
            "score": "[score from 1 to 5]"\\
        \},\\
        "overall": \{\\
            "reason": "[your rationale]",\\
            "score": "[score from 1 to 5]"\\
        \}\\
    \}\\

    Material 
    
    The following is provided for your evaluation: an edited video, encompassing both the script within the video and a series of video frames.
    
    Text Script: \\
    \{text\}
    
    Video Frames:\\
    Video Frames are shown below.
}
''
\end{quote} 

We compare the GPT-4 scores of our framework and three baseline models, which are obtained by GPT-4 scoring the text and video frames of the created videos in four aspects. In addition, we also compute the average of these four scores for GPT-4. As shown in Table \ref{tab:gpt4_pad} and Table \ref{tab:gpt4_ucf101}, on the PAD and UCF101-DVC, our framework outperforms the three baseline models in all four aspects and their averages. This indicates that our framework excels in consistency, logicality, vividness, and overall compared to the three baseline models.

\begin{table}
\centering
\footnotesize
\setlength{\tabcolsep}{1.5pt}
\small{
\begin{tabular}{ccccc} 
\toprule 
\multicolumn{1}{c}{\textsc{Method}} & w/o Theme & w/o Blurred BG. & w/o ChatGPT & Ours \tabularnewline 
\midrule 
Consistency & 4.68 & 4.60 & 4.36 & \textbf{4.72} \tabularnewline 
Logicality & 4.64  & 4.48 & 4.36 & \textbf{4.72}  \tabularnewline 
Vividness  & 4.40 & 4.32 & 4.04 & \textbf{4.52} \tabularnewline 
Overall  & 4.48 & 4.44 & 4.28 & \textbf{4.56}  \tabularnewline 
Average  & 4.55 & 4.46 & 4.26 & \textbf{4.63}  \tabularnewline 
\bottomrule
\end{tabular}}
\caption{
Quantitative comparison of GPT-4 scores between our framework and three baseline models on the PAD. The ``Average'' represents the average score across the four  aspects. Larger numbers in the table indicate superior performance or preference.
} \label{tab:gpt4_pad}
\vspace{-0.15in}
\end{table}

\begin{table}
\centering
\footnotesize
\setlength{\tabcolsep}{1.5pt}
\small{
\begin{tabular}{ccccc} 
\toprule 
\multicolumn{1}{c}{\textsc{Method}} & w/o Theme & w/o Blurred BG. & w/o ChatGPT & Ours \tabularnewline 
\midrule 
Consistency & 4.25 & 4.31 & 3.34 & \textbf{4.50} \tabularnewline 
Logicality & 4.20  & 4.33 & 2.90 & \textbf{4.48}  \tabularnewline 
Vividness  & 3.66 & 3.71 & 2.58 & \textbf{4.03} \tabularnewline 
Overall  & 3.64 & 3.69 & 2.70 & \textbf{3.93}  \tabularnewline 
Average  & 3.94 & 4.01 & 2.88 & \textbf{4.24}  \tabularnewline 
\bottomrule
\end{tabular}}
\caption{Quantitative comparison of GPT-4 scores between our framework and three baseline models on the UCF101-DVC. The ``Average'' represents the average score across the four  aspects. Larger numbers in the table indicate superior performance or preference.} \label{tab:gpt4_ucf101}
\vspace{-0.15in}
\end{table}

\section{Additional Qualitative Results}

\begin{figure*}
    \centering
    \includegraphics[height=\textheight]{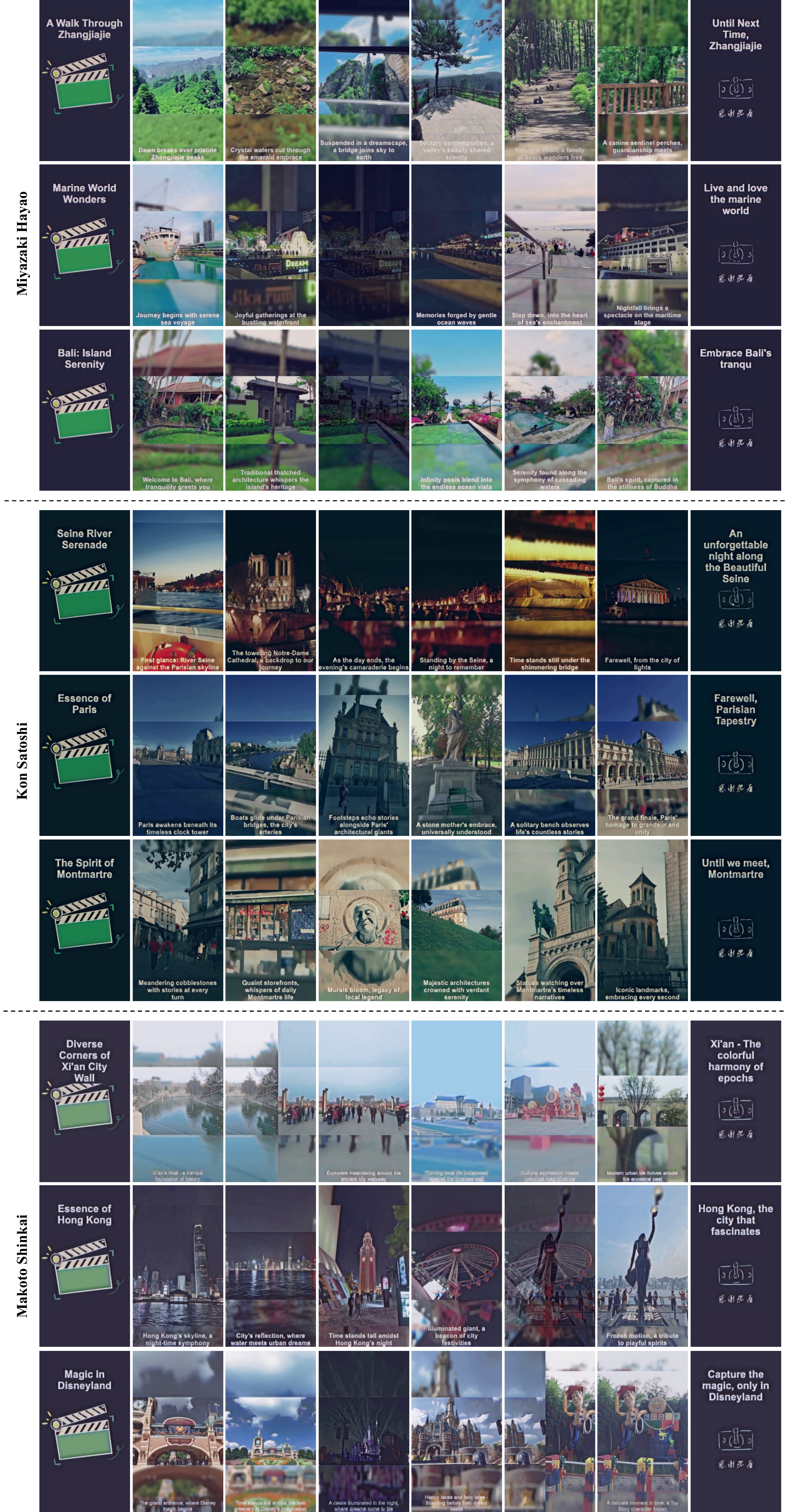}
    \caption{Results of style transfer with three animated styles on the Personal Album Dataset.}
    \label{fig:PAD_style}
\end{figure*}

\begin{figure*}
    \centering
    \includegraphics[height=\textheight]{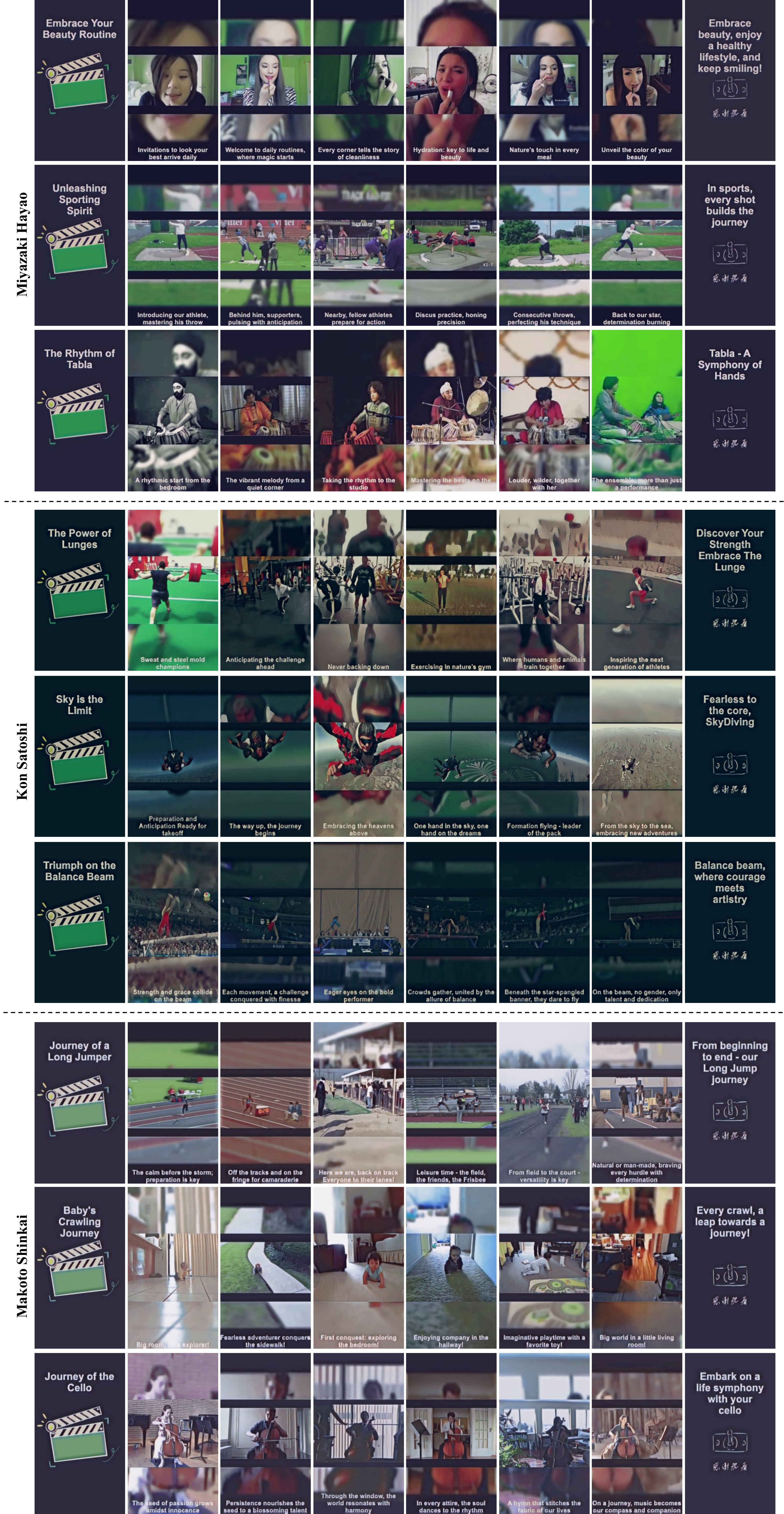}
    \caption{Results of style transfer with three animated styles on the UCF101-DVC Dataset.}
    \label{fig:UCF101_style}
\end{figure*}

\begin{figure*}
    \centering
    \includegraphics[height=\textheight]{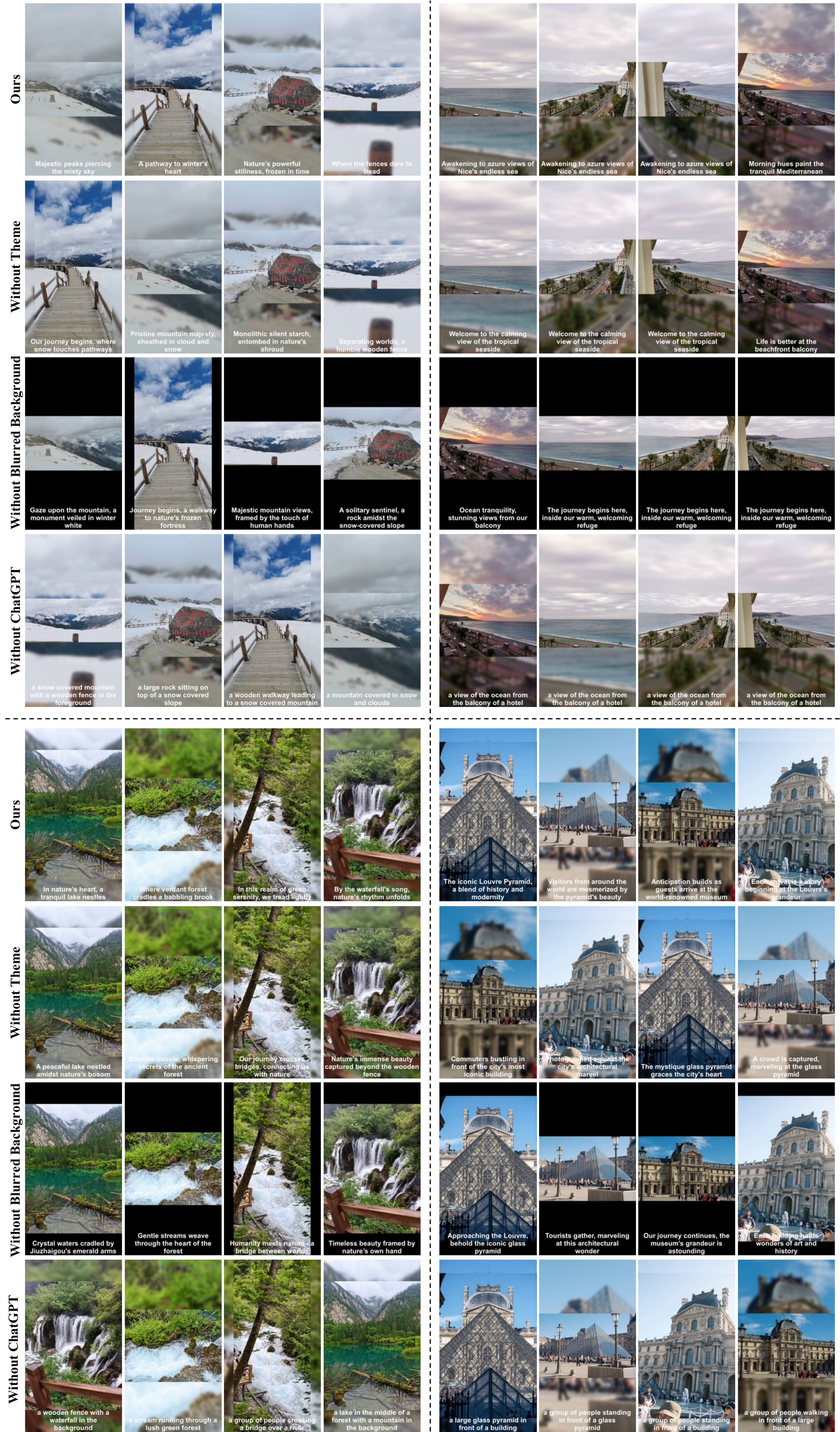}
    \caption{Comparison results between our framework and three baseline models on the Personal Album Dataset.}
    \label{fig:PAD_res_supp}
\end{figure*}

\begin{figure*}
    \centering
    \includegraphics[height=\textheight]{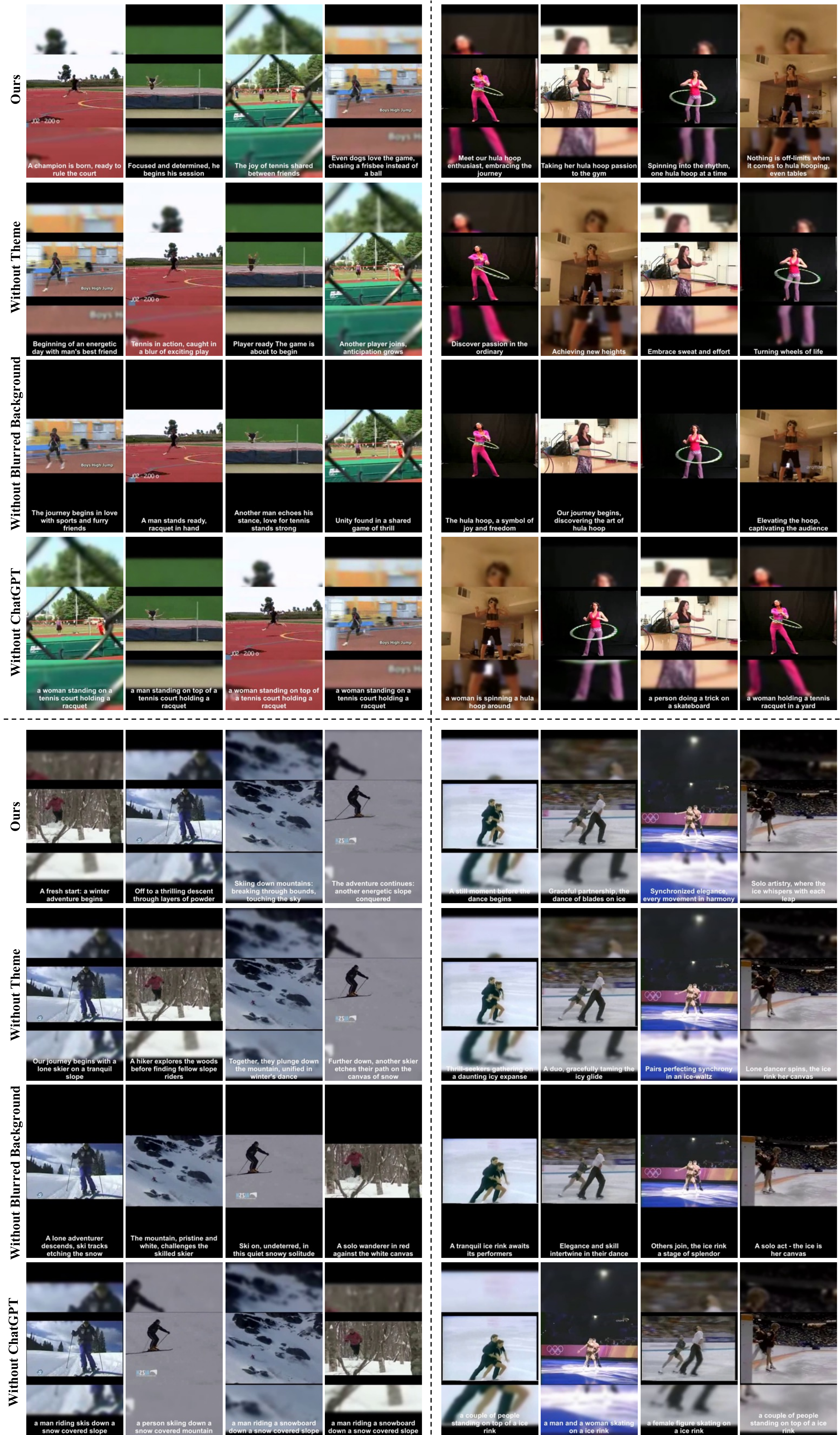}
    \caption{Comparison results between our framework and three baseline models on the UCF101-DVC Dataset.}
    \label{fig:UCF101_res_supp}
\end{figure*}

As shown in Figure \ref{fig:PAD_style}, we present the results of style transfer of our framework on the PAD across three animated styles.  At the same time, illustrated in Figure \ref{fig:UCF101_style} is the results of style transfer of our framework on the UCF101-DVC Dataset across three animated styles. These results demonstrate the capability of our framework to generate diverse, aesthetically pleasing, and coherent storytelling videos. Furthermore, Figure \ref{fig:PAD_res_supp} provides a comparative analysis between our framework and three baseline models on PAD, while Figure \ref{fig:UCF101_res_supp} provides the corresponding comparisons on the UCF101-DVC dataset. The results indicate that our framework produces captions closely aligned with the video content, forming a coherent story. The results of ``Without Theme'' suggest that user-provided themes contribute to captions better aligning with the visual content of the video. The results of ``Without Blurred Background'' highlight the significant enhancement of video aesthetics with the addition of Gaussian-blurred backgrounds. Lastly, the results of ``Without ChatGPT''  demonstrate the role of ChatGPT in generating more coherent and storytelling captions. These results offer a comprehensive view of the qualitative aspects of our proposed framework, further supporting its efficacy in Dynamic Visual Composition tasks.

\end{document}